\documentclass[review]{elsarticle}

\usepackage{hyperref}

\usepackage{amsfonts, amsmath, amsthm, amssymb}
\usepackage{siunitx}

\usepackage[utf8]{inputenc} 
\usepackage[T1]{fontenc}    
\usepackage{hyperref}       
\usepackage{url}            
\usepackage{booktabs}       
\usepackage{amsfonts}       
\usepackage{nicefrac}       
\usepackage{microtype}      
\usepackage{xcolor}         
\usepackage{rotfloat}

\usepackage{ltablex}

\usepackage{multirow}

\usepackage{graphicx}

\usepackage{caption}
\usepackage{subcaption}
\usepackage{float}

\usepackage{mathtools}
\usepackage{url}
\usepackage{wrapfig}

\newcommand{\dpcgans}{DP-CGANS}

\begin{document}

\begin{frontmatter}

\title{Improving Correlation Capture in Generating Imbalanced Data using Differentially Private Conditional GANs}

\author[idsaddress]{Chang Sun\corref{mycorrespondingauthor}}
\cortext[mycorrespondingauthor]{Corresponding author}
\ead{chang.sun@maastrichtuniversity.nl}

\author[bissaddress]{Johan van Soest}

\author[idsaddress]{Michel Dumontier}

\address[idsaddress]{Institute of Data Science, Maastricht University, Maastricht, The Netherlands}
\address[bissaddress]{Brightlands Institute of Smart Society, Maastricht University, Heerlen, The Netherlands}

\begin{abstract}
 Despite the remarkable success of Generative Adversarial Networks (GANs) on text, images, and videos, generating high-quality tabular data is still under development owing to some unique challenges such as capturing dependencies in imbalanced data, optimizing the quality of synthetic patient data while preserving privacy. In this paper, we propose DP-CGANS, a differentially private conditional GAN framework consisting of data transformation, sampling, conditioning, and networks training to generate realistic and privacy-preserving tabular data. DP-CGANS distinguishes categorical and continuous variables and transforms them to latent space separately. Then, we structure a conditional vector as an additional input to not only presents the minority class in the imbalanced data, but also capture the dependency between variables. We inject statistical noise to the gradients in the networking training process of DP-CGANS to provide a differential privacy guarantee. We extensively evaluate our model with state-of-the-art generative models on three public datasets and two real-world personal health datasets in terms of statistical similarity, machine learning performance, and privacy measurement. We demonstrate that our model outperforms other comparable models, especially in capturing dependency between variables. Finally, we present the balance between data utility and privacy in synthetic data generation considering the different data structure and characteristics of real-world datasets such as imbalance variables, abnormal distributions, and sparsity of data. 
\end{abstract}


\end{frontmatter}


\section{Introduction}
\label{intro}
Data from individuals such as personal health or behavior data have proven to be highly valuable for the scientific community~\cite{nass2009value,kalkman_patients_2022}. This data is sensitive and requires special attention and protection. Due to disclosure limitations and legal requirements, such data is not always accessible for the scientific community~\cite{resnik2006openness,european_commission_directorate_general_for_communications_networks_content_and_technology_study_2021,walonoski_synthea_2018}. Even if some data is accessible by request, researchers need to invest enormous time and effort in the requesting process which may take months or years without knowing if the data is sufficiently useful for the research studies. This can cause a severe delay and inordinate costs for research projects~\cite{lugg2018challenges,dattani_accessing_2013}. To mitigate this issue, we propose to generate and use synthetic data that is structurally and statistically similar to real data at the population level (i.e., distributions of single variables, correlations between variables), and machine learning utility level (i.e., the analysis results on synthetic data are comparable to the results on real data). An example is that data parties provide researchers with realistic synthetic data to construct machine learning models. Afterwards, the built models are sent to data parties to be executed on the source data and only return the results to the researchers. The realistic synthetic data offers the possibility for researchers to i) assess whether the data are relevant for their studies and ii) obtain statistically valid insights without access to the underlying data or before starting the data requesting process. 


One of the recent promising method to generate synthetic data is Generative Adversarial Network (GAN) which has been successfully developed to generate synthetic image, text, music, health and financial data with promising performance~\cite{briot2021artificial,Wiese2020,baowaly2019synthesizing,kurup2021evolution,Zhang2017GAN}. However, using GANs to generate tabular data poses exclusive challenges~\cite{xu_modeling_2019,zhao_ctab_gan_2021}. One challenge is that the correlations and dependencies among imbalanced variables are typically not well-preserved in the generated synthetic data. It is crucial to transfer such information from real data to synthetic data in many domains such as healthcare and social sciences. For instance, we would expect the preservation of a positive relationship between daily physical activity and mobility in synthetically generated health data. Another challenge in working with personal data lies in the possibility of using GANs to security attacks that accurately reveal missing characteristics of real individuals, which could compromise their privacy ~\cite{shokri_2017,hayes2019logan,fredrikson_2015}. Optimizing the trade-off between the privacy of the source data and the quality of the synthetic data remains an open challenge~\cite{goncalves_generation_2020,howe_synthetic_2017}. 

In this paper, we propose a DP-CGANS framework (Differentially Private - Conditional Generative Adversarial NetworkS), consisting of four components including transformation, sampling, conditioning, and networking training with differential privacy, to generate realistic and privacy-preserving synthetic data. DP-CGANS constructs conditional vectors and an extra penalty to enforce the generator to captures the under-represented classes in the imbalanced variables and simulate the correlations and dependencies between these imbalanced variables. To motivate the model to generate diverse and representative synthetic data, we apply Wasserstein distances with gradient penalty and then group the training samples to the discriminator. Finally, we provide a privacy guarantee through  a differential privacy approach that injects Gaussian noise to the penalty gradients in the training process. Under a certain differential privacy threshold, DP-CGANS prevents the synthetic data from leaking sensitive information originating in the source data. We conduct experiments on three public datasets and two real-life personal health data comparing with the other three state-of-the-art generative models. The performance is evaluated on statistical similarity, machine learning performance, and privacy risks in attribute and identity disclosure under varying differential privacy budgets. Results indicate that DP-CGANS outperforms other comparable models for most datasets and captures the most dependencies between imbalanced variables. We observe the offer a trade-off between data utility and privacy in synthetic data generation. 


This paper is structured as follows: Section 2 describes background knowledge on conditional GANs and differential privacy, followed by the state-of-art methods. Section 3 elaborates on our proposed methods and supporting theories. Section 4 presents a set of experiments and results on five datasets, followed by findings and discussion in Section 5. Finally, we conclude the study in section 6.

\section{Related Work}
\label{relatedwork}
Generative Adversarial Network (GAN) contains two neural networks - a generator and a discriminator competing with each other. The generator aims to create realistic synthetic data points that cannot be indistinguished by the discriminator, while the discriminator is trained to accurately classify real and synthetic data created by the generator. Unique challenges of using GANs to generate tabular data have been widely recognized such as modeling data with mixed types (categorical and continuous), preventing model collapse, handling imbalanced variables, and capture the dependencies among variables ~\cite{xu_modeling_2019,goodfellow2014generative}. Several variants of GANs have been proposed to overcome these challenges~\cite{goncalves_generation_2020}. MedGAN~\cite{choi2017generating} transforms the binary and discrete variables to a continuous space by combining an auto-encoder with a GAN. MedGAN is one of the earliest GAN variants to generate synthetic Electronic Health Records. It handles binary and continuous variables in separate models but not multi-categorical variables. TableGAN~\cite{park2018data} adds a third neural network as a classifier in addition to the generator and the discriminator to increase the semantic integrity of the synthetic data. TableGAN has good performance on handling discrete and continuous variables but suffers from model collapse with categorical data. 

\subsection{Handling Mode Collapse}
Model collapse occurs when the generator discovers some data points that are classified as real data by the discriminator with high confidence and then replicates them to all the data points. In this case, the discriminator fails to provide useful gradients to the generator anymore. To address this challenge, Martin et al.~\cite{arjovsky_wasserstein_2017} proposed a WGAN which used Wasserstein distance to measure the minimal cost of transforming random data points from an arbitrary distribution into the other target distribution. Further, Ishaan et al.~\cite{gulrajani_improved_2017} introduced a gradient penalty to penalize the discriminator, called WGAN-GP, which stabilized WGAN training and better-prevented vanishing gradients. In addition to WGANs which adjust the objective functions, PacGAN~\cite{lin_pacgan_2018} restructures the discriminator from mapping one data sample to a class (real or synthetic) to mapping a set of independent samples to a class. The packed discriminator can effectively detect mode collapse when there is a lack of diversity in a set of data samples. The objective function of WGAN-GP is constructed as equation~\ref{eq:gan_3}. The coefficient $\lambda$ is defined as the weight of gradient penalty term in the training. $P_{\widehat{x}}$ is the distribution uniformly sampled between the real ($P_r$) and generator model distribution ($P_g$).  

\begin{equation}
\label{eq:gan_3}
L = \underbrace{\mathit{\mathbb{E}}_{\widetilde{x} \sim P_g}[D(\widetilde{x})] - \mathit{\mathbb{E}}_{x \sim P_r}[D(x)]}_{WGAN Loss} + \underbrace{\lambda \mathit{\mathbb{E}}_{\widehat{x}\sim P_{\widehat{x}}}[(\parallel \bigtriangledown_{\widehat{x}} \space D(\widehat{x})\parallel_2 -1)^2]}_{Gradient Penalty}
\end{equation}

\subsection{Handling Imbalanced Data}
When generating imbalanced data, the major category is likely to dominate the training of the discriminator so that the discriminator fails to detect the absence of the minor category. Conditional GAN (CGAN) append an additional vector to the input of the generator and discriminator to address this concern. Engelmann and Lessmann~\cite{engelmann_conditional_2021} proposed CW-GAN, a CGAN using the WGAN-GP objective function, as an oversampling method for imbalanced datasets with both continuous and categorical variables. With a similar goal, CTGAN~\cite{xu_modeling_2019} invents a training-by-sampling method to handle imbalanced categorical variables in addition to the conditional vector. Based on CTGAN and TableGAN, CTAB-GAN~\cite{zhao_ctab_gan_2021} combines two frameworks to solve the challenges in industrial datasets such as variables with mixed data types and long-tail distributions. 

\subsection{Handling Privacy Concerns}

GANs could elicit privacy concerns when the training data is personal and/or sensitive~\cite{rosenblatt_differentially_2020,kunar_effective_2021}. To protect the source data from malicious privacy attacks, recent work shows the promising application of combining Differential Privacy (DP) into GANs~\cite{torfi_differentially_2020,torkzadehmahani_dp_cgan_2020,xie_differentially_2018}. DP uses a solid mathematical formulation to measure the privacy and provide theoretical privacy guarantees by typically adding noise when training the models~\cite{dwork2008differential,Mironov2017RnyiDP}. A model is considered to be ($\varepsilon$, $\delta$) - differentially private if for any two datasets $D$ and $D^\prime$ differing in a single data point and for any subset of outputs $S$: 

\begin{equation}
\label{eq:gan_4}
\mathbb{P}(M_p(D)\in S) \leqslant e^\varepsilon \cdot \mathbb{P}(M_p(D\prime)\in S) + \delta
\end{equation}

where $M_p(D)$ and $M_p(D^\prime)$ are the outputs of the model for input datasets $D$ and $D^\prime$, $\mathbb{P}$ is the randomness of the noise, $\varepsilon$ reflects the privacy level. A small $\varepsilon$ ($\leqslant$ 1.0) indicates the small difference of model's output probabilities on $D$ and $D^\prime$ which results in a high privacy guarantee. Differential privacy can protect the participation of individual data points in the datasets, which means replacing or removing one data point (data instance) with another one will not make an observable change in the analysis results. Xie et al.~\cite{xie_differentially_2018} added noise on the gradient of Wasserstein distance during the discriminator training, while Chen et al.~\cite{chen2020gs} uses WGAN-GP framework and inject noises in the generator training. However, both models were designed and experimented on image data.


\section{Methodology}
\begin{figure}[hbt!]
    \centering
    \includegraphics[scale=0.11]{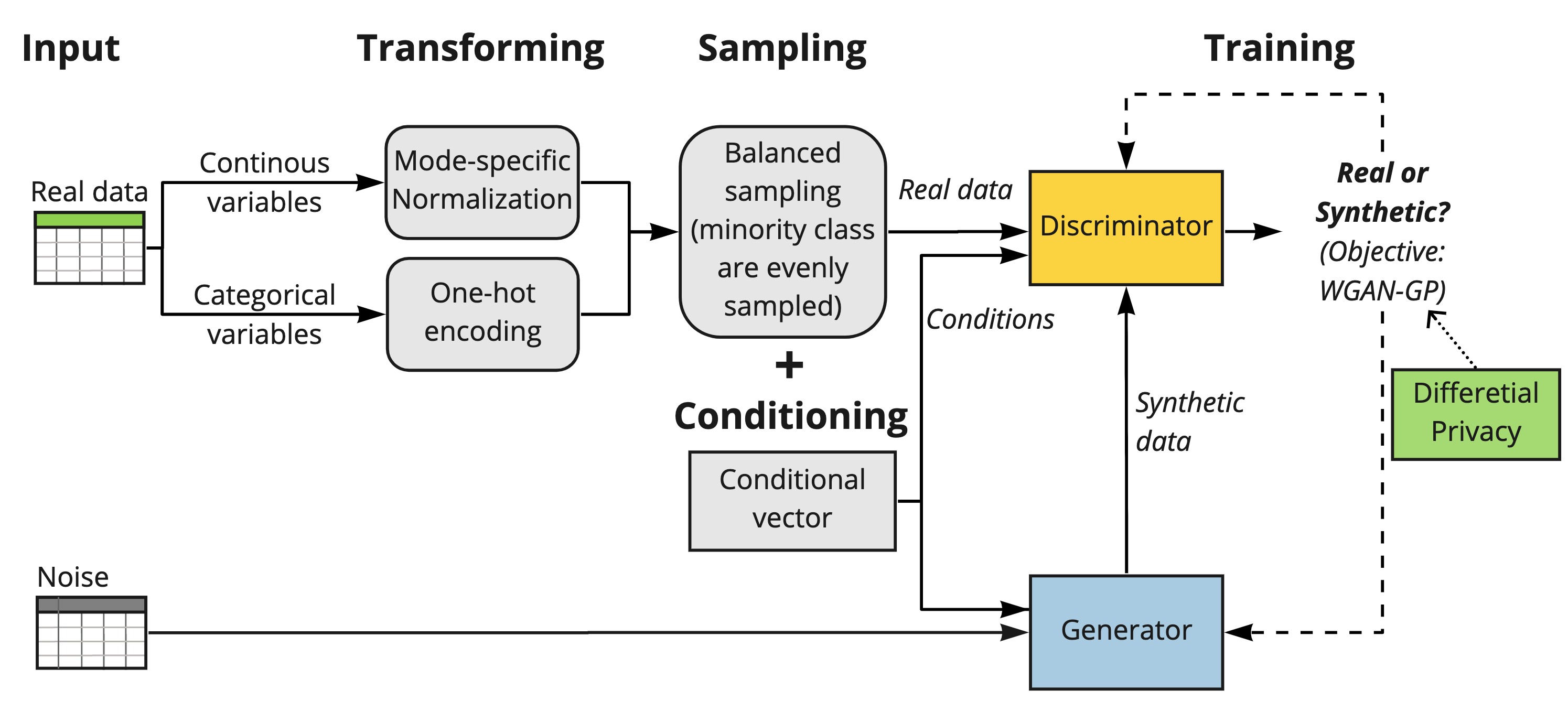}
    \caption{\textcolor{black}{The overall structure of \dpcgans Framework}}
    \label{cha6fig:workflow}
\end{figure}

 The development of DP-CGANS is based on the strengths of prior studies including~\cite{walia_synthesising_2020,xu_modeling_2019,xie_differentially_2018} and further extended to address the remaining challenges including better handling imbalanced variables, capturing the correlations and dependencies between variables, and adding differential privacy guarantee. The overall framework of DP-CGANS is illustrated in Figure~\ref{cha6fig:workflow} including four main steps which are transformation, sampling, conditioning, and training. DP-CGANS separates input data to categorical and continuous variables to apply different transformation methods and activation functions. Categorical variables are encoded using one-hot encoding method, while continuous variables are transformed using mode-specific normalization proposed by Xu et al.~\cite{xu_modeling_2019}. In real-world datasets, continuous variables commonly have multimode distribution such as heights (body length) of males and females. Instead of forcing the values of continuous variables to [-1, 1] using a min-max transformation, mode-specific normalization estimates the number of the distributions of continuous variables by a variational Gaussian mixture model with Dirichlet Process. The values of each continuous variable are normalized according to its estimated distributions. After training, the synthetic data produced by the generator is inversely transformed back to the original scales. 

\subsection{Conditional vectors and data samples for training }
The key process in DP-CGANS to capture the dependencies between imbalanced variables is sampling and conditioning. Our method is leveraged from the training-by-sampling method and the design of the conditional generator in~\cite{xu_modeling_2019}. The primary idea is to encode the values of each categorical variable and present them as an additional conditional vector to train the generator and discriminator. The input training data is resampled based on this conditional vector to ensure the minority category values can also be observed in the training. 

\begin{figure}[hbt!]
    \centering
    \includegraphics[scale=0.1]{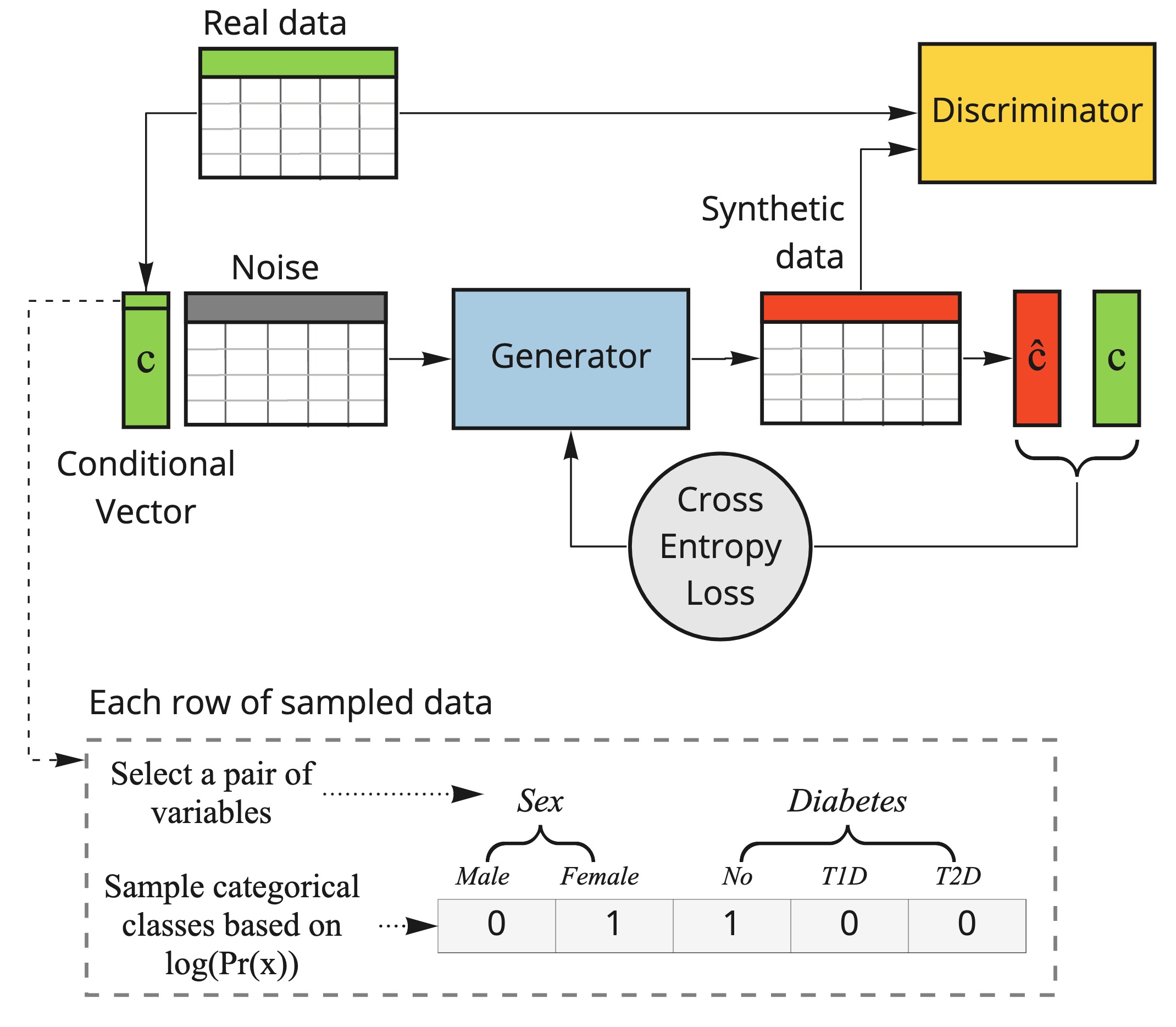}
    \caption{\textcolor{black}{Construction of the conditional vector and generator training.}}
    \label{cha6fig:condvec}
\end{figure}

However, the existing approach and its variants~\cite{zhao_ctab_gan_2021,moon_conditional_2020} treat each variable independently, thereby potentially losing the dependencies between variables in the synthetic data. DP-CGANS addresses this issue by conditioning the generator with an extensive vector representing the dependencies between variables.  Figure~\ref{cha6fig:condvec} shows the construction of the conditional vectors in DP-CGANS. After transformation, each value in the categorical variables is encoded into a one-hot vector. For each row in the sampled data, we randomly select and pair two categorical variables with equal probability. In the example, they are variables of Sex and Diabetes. Then, the probability mass of every possible combination of categorical values is calculated from these two variables and one pair out of all possible combinations is sampled. The sampled values are Female and No diabetes and are represented as 01 100 in the example. The occurrence of two rare categorical values is relatively low and difficult to capture in this case. The sampling is based on the logarithm of the probability which increases the chance of picking up the dependency between two rare categorical values. 

\subsection{Network structure and training method }
\begin{figure}[hbt!]
    \centering
    \includegraphics[scale=0.1]{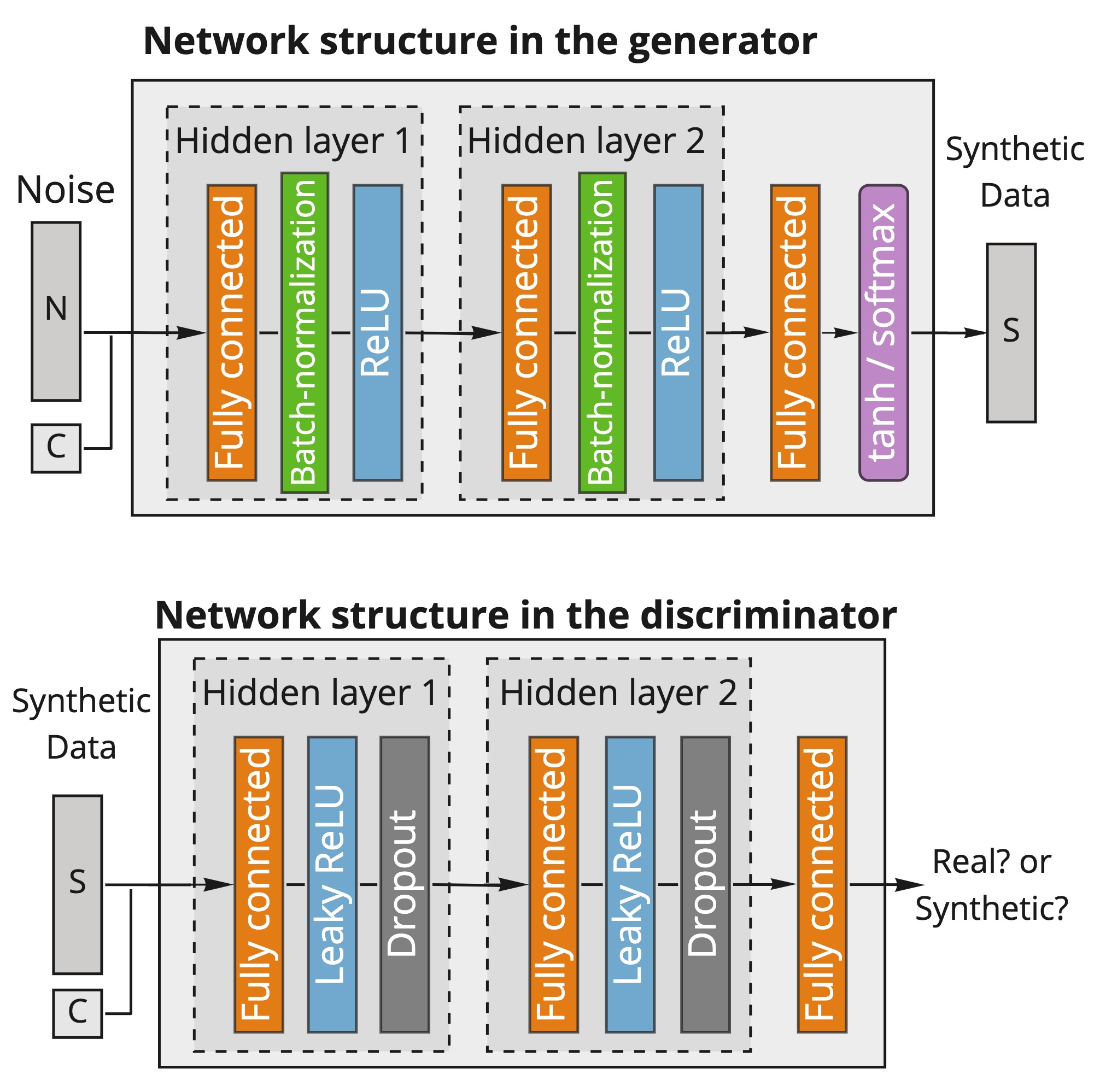}
    \caption{\textcolor{black}{Network Structure of the generator and the discriminator.}}
    \label{cha6fig:network}
\end{figure}

Figure~\ref{cha6fig:network} illustrates the network structures of the generator and the discriminator of DP-CGANS. The \textbf{generator} uses a fully connected network with two hidden layers. Each hidden layer applies the batch-normalization and Relu activation function for efficiency and stability purposes to address the vanishing gradient and data sparse problems. To generate the mix of categorical and continuous features, \textit{tanh} and \textit{softmax} activation functions are applied on the output layer~\cite{xu_synthesizing_2020}.

The generator is trained to produce more realistic synthetic data by learning the loss based on the discriminator's classification of the real and synthetic data. Figure~\ref{cha6fig:condvec} shows the generator training process in DP-CGANS. In addition to learning from the loss output from the discriminator, the generator takes an extra penalty to present the variables and values which are sampled in the conditional vector ($\widehat{C}$) and to maximally mimic the conditional vector from the real data (C). As the conditional vector includes multiple variables to capture their dependency, we introduced a Binary Cross-Entropy Loss combined with a Sigmoid layer to penalize the generator loss in DP-CGANS. 

The \textbf{discriminator} is a fully connected network with two hidden layers. The hidden layers apply Leaky Relu functions which can handle negative input values and has better performance than Relu in the discriminator and dropout on each layer~\cite{xu2015empirical,maas2013rectifier}. To mitigate mode collapse, the discriminator is constructed following the PacGAN framework, which is an augmented discriminator mapping 10 samples to a single class. DP-CGANS applies the objective function in equation (3) following the WGAN-GP (Wasserstein GAN with gradient penalty) structure with gradient penalty coefficient 10. The more the discriminator is trained, the more useful gradient of the Wasserstein will be obtained. We run 5 iterations of the discriminator in each generator iteration. Lastly, the discriminator and generator both use Adam optimization with learning rate ($\alpha$) \num{1e-4}, the exponential decay rate for the first and second moment estimates ($\beta_1$, $\beta_2$) 0.5 and 0.99. 

\subsection{Differential Privacy in \dpcgans}
DP-CGANS takes real data as input to train the discriminator. To protect these data, we inject Gaussian noise to the penalty gradient of the Wasserstein distance while training the discriminator~\cite{xie_differentially_2018}. A post-processing property of differential privacy has been shown that operations after a differentially private output will not violate the privacy~\cite{Cynthia_2014}. Therefore, privatizing the discriminator can impose the generator to become differentially private in that the generator is trained based on the differentially private discriminator's output~\cite{rosenblatt_differentially_2020,xie_differentially_2018}. 

In each iteration, the discriminator calculates the gradients of loss to optimize its training objective. We clip the gradients to [$-C_p$, $+C_p$ ] where $C_p$ is the parameter clip constant and inject Gaussian noise $(N(0, \sigma^2(C_g)^2I))$ to the clipped gradient where $\sigma$ is the noise scale, $C_g$ is the bound on the gradient of Wasserstein distance. To monitor the spent privacy budget ($\varepsilon$, $\delta$), the model tracks and checks the privacy budget every time the noise is added to the gradient. Different from existing methods using moment accountant technique, we applied R\'enyi Differential Privacy (RDP) Accountant~\cite{Mironov2017RnyiDP} which calculates a tighter estimation of privacy budget. At every iteration step, the privacy budget is bounded and accumulated. When the total privacy budget exceeds the initial target, the training process will be terminated and DP-CGANS is able to generate differentially private synthetic data.

\section{Experiments and results}
The experiment includes three public datasets that are commonly used by the machine learning community from UCI Machine Learning Repository~\cite{blake1998uci} and two real-world personal health datasets (Table~\ref{cha6tab:datatable}). All  datasets contain multiple data types – continuous, binary, and categorical. The Adult dataset~\cite{adultdataset} and Census dataset~\cite{censusdataset} contain socio-economic data from individuals, while Census has a majority of categorical variables. Intrusion dataset~\cite{intrusiondataset} is about network intrusion detections with most continuous variables. To reduce the computation time, 12k rows and 123k rows of data were randomly sampled from Census and Intrusion datasets in a stratified way with respect to the target variables. The Diabete dataset is requested from the Maastricht Study, an observational prospective population-based cohort study focusing on Type 2 Diabetes~\cite{schram2014maastricht}. The diabetes dataset includes demographic, socioeconomic, lifestyle, T2DM data of individuals. The cancer dataset is the clinical outcome data of non-small cell lung cancer (NSCLC) patients collected by the Maastro Clinic~\cite{aerts_decoding_2014,aerts_data_2019}. More details about the data collection of the two real-world datasets can be found in the appendix.


\begin{table}[]
\scriptsize
\centering
\caption{Description of experimented datasets. \#Cat represents the number of multi-class categorical variables, \#Con represents the number of continuous variables, and \#Bi represents the number of binary variables in the datasets.}
\label{cha6tab:datatable}
\begin{tabular}{|l|l|l|l|l|l|l|}
\hline
\textbf{Datasets} & \textbf{\#Rows} & \textbf{\#Cat} & \textbf{\#Con} & \textbf{\#Bi} & \textbf{Source} & \textbf{Access} \\ \hline
\textbf{Adult} & 30162 & 7 & 6 & 2 & UCI & Public \\ \hline
\textbf{Census} & 12000 & 31 & 7 & 3 & UCI & Public \\ \hline
\textbf{Intrusion} & 123000 & 4 & 20 & 0 & UCI & Public \\ \hline
\textbf{Diabetes} & 2257 & 8 & 10 & 6 & DMS & On request \\ \hline
\textbf{Cancer} & 365 & 5 & 2 & 2 & Maastro & On request \\ \hline
\end{tabular}
\end{table}

\subsection{Experiment Setting}
We compared the  performance of DP-CGANS with other three well-known GAN frameworks for generating tabular data - CTGAN~\cite{xu_modeling_2019}, MedGAN~\cite{choi2017generating}, and TableGAN~\cite{park2018data}. We applied the comprehensive benchmarking suite developed by Synthetic Data Gym Framework\footnote{SDGym Github Repository: https://github.com/sdv-dev/SDGym} where all these models were programmed in python using PyTorch library. We keep the original structure of their framework and use the same model parameters as they stated in their published studies. All models share the same number of epochs (\textit{2000}) and batch size (\textit{500}). We present the key hyperparameters of DP-CGANS for reproducibility purposes in Table~\ref{cha6tab:modelpara}. The experiments were conducted using one 32GB GPU (Nvidia DGX1 8x Tesla V100) in an OKD 4.6 cluster under the Data Science Research Infrastructure at Maastricht University\footnote{Data Science Research Infrastructure (DSRI): https://maastrichtu-ids.github.io/dsri-documentation/}.


\begin{table}[!htb]
\scriptsize
\centering
\caption{Key hyperparameters in DP-CGANS.}
\label{cha6tab:modelpara}
\begin{tabular}{llll}
\hline
\textbf{Step} & \textbf{Model} & \textbf{Hyperparameter} & \textbf{Value} \\ \hline
\multirow{5}{*}{\begin{tabular}[c]{@{}l@{}}Transfor-\\mation\end{tabular}} & \multirow{5}{*}{\begin{tabular}[c]{@{}l@{}}Gaussian \\ Mixture\end{tabular}} & Prior type for the weights' distribution & Dirichlet Process \\  
 &  & Max. num of Gaussian distribution & 10 \\ 
 &  & Weight concentration prior & \num{1e-3} \\  
 &  & Weight threshold & \num{1e-3} \\  
 &  & Num of mixture components & \textless{}=10 \\ \hline
\multirow{11}{*}{\begin{tabular}[c]{@{}l@{}}Network \\ training\end{tabular}} & - & Epochs & 2000 \\ 
 & - & Batch size & 500 \\  
 & Adam & Learning rate ($\alpha$) & \num{1e-4} \\  
 & PacGAN & Pac & 10 \\ 
 & WGAN-GP & Gradient penalty factor ($\lambda$) & 10 \\
 & Softmax & Non-negative scalar temperature ($\tau$) & 0.2 \\ 
 & LeakyReLU & Negative slope & 0.2 \\ 
 & Dropout & Probability of an element is 0 & 0.5 \\ \hline
\multirow{3}{*}{\begin{tabular}[c]{@{}l@{}}Differential \\ Privacy\end{tabular}} & - & Clip constant ($Cp$) & 0.01 \\ 
 & - & Probability of information leakage($\delta$) & \num{1e-5} \\ 
 & - & Privacy budget ($\epsilon$). & 0.1, 1, 10, 100, $\infty$ \\ \hline
\end{tabular}
\end{table}


\subsection{Evaluation Matrics}
A set of metrics are applied to comprehensively evaluate the performance of DP-CGANS and compared with other state-of-the-art models. The metrics are grouped to test the data utility of the synthetic data and measure the privacy cost of the generative model. The data utility metrics measure the statistical similarity between real and synthetic data and compare the machine learning performance. The privacy cost metrics measure how much information from the real data may be disclosed by the synthetic data and the generative models. An overview of evaluation metrics applied in this study is reported in Table~\ref{cha6tab:evastats}. 

\begin{table}[!b]
\scriptsize
\centering
\caption{An overview of evaluation metrics for synthetic data.}
\label{cha6tab:evastats}
\begin{tabular}{|l|l|l|l|}
\hline
\textbf{Metrics} & \textbf{Level} & \textbf{Method} & \textbf{Data Type} \\ \hline
\multirow{6}{*}{\begin{tabular}[c]{@{}l@{}}Statistic\\ similarity\end{tabular}} & \multirow{4}{*}{\begin{tabular}[c]{@{}l@{}}Single \\ variable\end{tabular}} & Chi Square (CS) & Categorical \\ \cline{3-4} 
 &  & Kolmogorov-Smirnov (KS) & Continuous \\ \cline{3-4} 
 &  & \multirow{2}{*}{KL Divergence} & Categorical \\ \cline{4-4} 
 &  &  & Continuous \\ \cline{2-4} 
 & \multirow{2}{*}{\begin{tabular}[c]{@{}l@{}}Variable \\ pairs\end{tabular}} & Pearson correlation & Continuous \\ \cline{3-4} 
 &  & Cramer's V coefficient & Categorical \\ \hline
\multirow{4}{*}{\begin{tabular}[c]{@{}l@{}}ML perfor \\ mance\end{tabular}} & \multirow{4}{*}{\begin{tabular}[c]{@{}l@{}}Whole \\ dataset\end{tabular}} & Logistic regression & - \\ \cline{3-4} 
 &  & Decision tree & - \\ \cline{3-4} 
 &  & Random forest (Adaboost) & - \\ \cline{3-4} 
 &  & Multi-layer perceptron & - \\ \hline
\multirow{4}{*}{\begin{tabular}[c]{@{}l@{}}Privacy \\ cost\end{tabular}} & \multirow{2}{*}{\begin{tabular}[c]{@{}l@{}}Identity\\ (Rows)\end{tabular}} & Hamming distance & Categorical \\ \cline{3-4} 
 &  & Euclidean distance & Continuous \\ \cline{2-4} 
 & \multirow{2}{*}{\begin{tabular}[c]{@{}l@{}}Attributes \\ (Columns)\end{tabular}} & Linear regression & Categorical \\ \cline{3-4} 
 &  & K-Nearest Neighbor & Continuous \\ \hline
\end{tabular}
\end{table}

\subsubsection{Data Utility Evaluation Metrics}

\textbf{Statistical Similarity}. We measured the statistical similarity by comparing the distribution of each variable independently and the correlation between variables. We include Kullback Leibler (KL) Divergence~\cite{hershey2007approximating}, Pearson's Chi-Square (CS) test~\cite{Pearson1900}, Kolmogorov Smirnov (KS) test~\cite{massey1951kolmogorov}, and pairwise correlation difference (PCD)~\cite{goncalves_generation_2020}. The KL divergence calculates the marginal probability mass functions (PMF) for each variable independently of the real and synthetic data and measures the similarity of the PMFs of the two variables. It is an information-theory based and asymmetric distance measurement to observe the information change between distributions before and after inferencing. We normalized the score to [0, 1] by calculating 1 / (1 + KL divergence) . When the distributions of two variables are similar, the normalized scores approach  1. The final score is the average of the scores of all measured variables in the data. We apply CS and KS statistical tests on categorical and continuous variables respectively. Different from KL divergence, which measures information loss from one distribution to another, CS and KS tests are null hypothesis statistical tests. CS test checks if the frequencies of categorical values in synthetic data match the frequencies in real data. KS test measures a symmetric distance between two empirical cumulative distributions of the continuous variables.

The difference of dependencies between each pair of variables is measured by the Pearson correlation matrices for continuous variables and Cramér's V Coefficient for categorical variables~\cite{frey2018sage}. Cramér's V Coefficient is based on Pearson's chi-square test to measure how strongly two categorical variables are associated. The difference score is scaled between 0 to 1. The smaller the score, the less difference between synthetic and real data.

\textbf{Machine Learning Performance}. The motivation of this study is to enable researchers to build their data analysis model based on synthetic data. The analysis of the synthetic data is expected to be the same or similar to the analysis of the real data. Therefore, the experimental datasets are split to training sets (75\%) and test sets (25\%). The training sets are fed into the generative models to produce the synthetic data. Then, a set of machine learning models including Logistic regression (LR), decision tree (DT), random forest (RF), and multilayer perceptron (MLP) models are trained on the real training data and generated synthetic data separately. Last, the trained machine learning models are evaluated on the real test data using AUC and F1 scores. The better and more realistic synthetic data is, the smaller the difference in its machine learning performance from the real data. 

\subsubsection{Privacy Cost Evaluation Metrics}
The privacy metrics cover two main classes of information disclosure that may happen in the synthetic data – identity disclosure and attribute disclosure~\cite{goncalves_generation_2020}. \textbf{Identity disclosure} means an attacker can exactly identify an individual (data sample) in the training data, which can be understood as if we can find one or more synthetic data with a certain distance to a real data sample which is used to generate the synthetic data~\cite{choi2017generating}. Hamming distance for the categorical variables and Euclidean distance for the continuous variables are calculated on each sample from the synthetic dataset. The attacker may identify the data sample which is indeed used for training (True Positive, TP), identify the sample but the sample is not used for training (FP), correctly identify the sample which is indeed not used for training (TN), wrongly identify the sample which is not used for training (False Negative, FN). The final identity disclosure is measured using the precision and recall of the above scores.

\textbf{Attribute disclosure} can be interpreted as if an attacker can predict the original values of the synthesized variables (sensitive variables) from an individual level based on some other variables of the real data that are known to the attacker (known variables). We observe the average posterior probabilities of the attacker correctly predicting the sensitive variables on the real test data. The risk of attribute disclosure is affected by the number of known variables from the source data, the size of the synthetic data, and the attack model setting. A linear regression model is applied to the continuous variables, and a K-nearest-neighboring model is for the categorical variables. We experiment on different sets of known variables to predict other original variables on the same size of the synthetic data from different generative models.

\section{Results and discussion}
This section presents the experiment results (Jan 2022) of DP-CGANS, CTGAN, MedGAN, and TableGAN on five datasets. Each experiment was conducted 3 times and the results are the average of them. Then, we present the model performance of DP-CGANS using different privacy budgets on Adult and Diabetes datasets. Then, we describe the limitations of DP-CGANS based on our observations. 

\subsection{Statistical Similarity}
Evaluation results of statistical similarity are presented in Table~\ref{cha6tab:statsresults}. KL divergence, CS, and KS tests measure the similarity of each individual variable independently (the higher the score, the more similar between synthetic and real data), while Cramér's V coefficient and Pearson correlation measure the difference of dependencies between a pair of variables (the lower the score, the more dependencies in real data captured by synthetic data). DP-CGANS outperforms other models in the CS, KS, and KL Divergence test on categorical variables in most datasets. The conditional vector and the additional penalty in the generator of DP-CGANS successfully capture the underrepresented categories and the dependencies between them. CTGAN performs similarly to DP-CGANS because of its conditional GAN structure and sampling method. Both models can handle the datasets with imbalanced variables better than MedGAN and TableGAN. 

\begin{figure}[hbt!]
    \centering
    \includegraphics[scale=0.15]{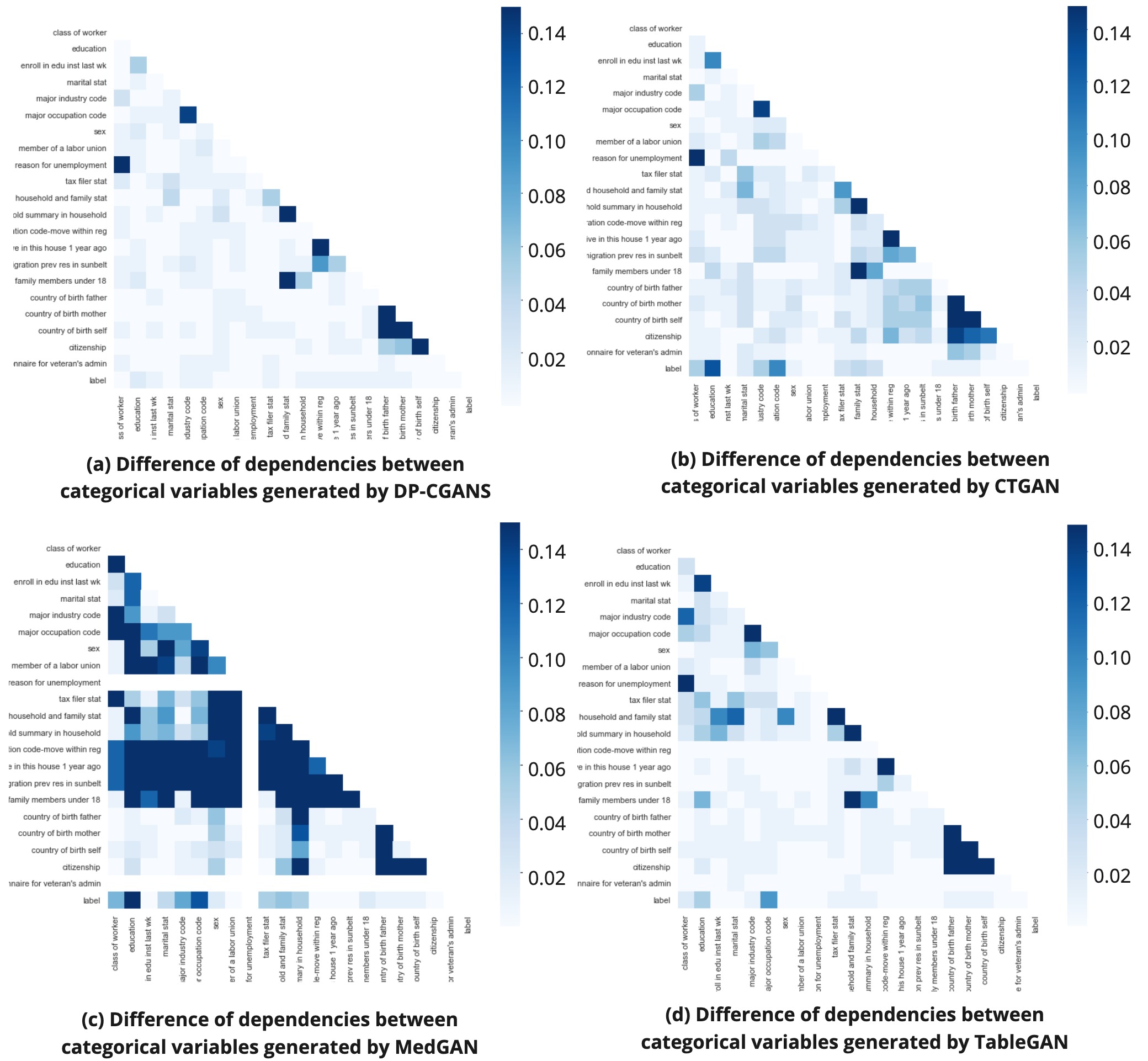}
    \caption{\textcolor{black}{Differences of dependencies between categorical variables from real to synthetic data generated by different models. The darker the cell, the greater the difference in dependence between two variables between real and synthetic data.}}
    \label{cha6fig:dependency}
\end{figure}

The results of Cramér's V Coefficient and Pearson correlation show that DP-CGANS is outstanding in simulating the dependencies and correlations between variables. Figure~\ref{cha6fig:dependency} shows the differences of dependencies between categorical variables from the Census dataset and the synthetic data generated by different models. The darker the blue of the cell, the greater the difference in dependence between two variables in the real and synthetic data. DP-CGANS simulates the most dependencies between variables followed by TableGAN and CTGAN, while MedGAN fails to transfer the most of dependencies. DP-CGANS outperforms TableGAN on the variables that have multiple major classes and many different minor classes. The reason is the additional penalty in the generator and the sampling method of training enable DP-CGANS to transfer the underrepresented dependencies of the minor classes in the imbalanced variables. DP-CGANS outperforms CTGAN in the variables that have one or two extreme dominant classes and several minor classes. The reason is that the construction of the conditional vector of DP-CGANS aims to capture the dependencies between imbalanced variables, but this is not presented in CTGAN. 

\begin{scriptsize}
\begin{longtable}[c]{llllll}
\caption{Results of measuring statistical similarity between real and synthetic data.}
\label{cha6tab:statsresults}\\
\toprule
\multicolumn{1}{l}{\textbf{}} & \multicolumn{1}{l}{Adult} & \multicolumn{1}{l}{Census} & \multicolumn{1}{l}{Intrusion} & \multicolumn{1}{l}{Diabetes} & Cancer \\ \hline
\endfirsthead
\endhead
\multicolumn{6}{c}{\textbf{KL Divergence (Categorical)}} \\ 
\multicolumn{1}{l}{DP-CGANS} & \multicolumn{1}{l}{\textbf{0.921}} & \multicolumn{1}{l}{\textbf{0.933}} & \multicolumn{1}{l}{\textbf{0.737}} & \multicolumn{1}{l}{\textbf{0.982}} & \textbf{0.918} \\ 
\multicolumn{1}{l}{CTGAN} & \multicolumn{1}{l}{0.894} & \multicolumn{1}{l}{0.834} & \multicolumn{1}{l}{0.708} & \multicolumn{1}{l}{0.957} & 0.881 \\ 
\multicolumn{1}{l}{MedGAN} & \multicolumn{1}{l}{0.785} & \multicolumn{1}{l}{0.746} & \multicolumn{1}{l}{0.605} & \multicolumn{1}{l}{0.905} & 0.620 \\ 
\multicolumn{1}{l}{TableGAN} & \multicolumn{1}{l}{0.746} & \multicolumn{1}{l}{0.856} & \multicolumn{1}{l}{0.603} & \multicolumn{1}{l}{0.941} & 0.814 \\ \hline
\multicolumn{6}{c}{\textbf{KL Divergence (Continuous)}} \\ \hline
\multicolumn{1}{l}{\textit{DP-CGANS}} & \multicolumn{1}{l}{0.887} & \multicolumn{1}{l}{\textbf{0.828}} & \multicolumn{1}{l}{0.906} & \multicolumn{1}{l}{0.801} & 0.552 \\ 
\multicolumn{1}{l}{\textit{CTGAN}} & \multicolumn{1}{l}{\textbf{0.929}} & \multicolumn{1}{l}{0.791} & \multicolumn{1}{l}{\textbf{0.922}} & \multicolumn{1}{l}{0.736} & 0.560 \\ 
\multicolumn{1}{l}{\textit{MedGAN}} & \multicolumn{1}{l}{0.115} & \multicolumn{1}{l}{0.083} & \multicolumn{1}{l}{0.198} & \multicolumn{1}{l}{0.184} & 0.180 \\ 
\multicolumn{1}{l}{\textit{TableGAN}} & \multicolumn{1}{l}{0.752} & \multicolumn{1}{l}{0.460} & \multicolumn{1}{l}{0.820} & \multicolumn{1}{l}{\textbf{0.860}} & \textbf{0.565} \\ \hline
\multicolumn{6}{c}{\textbf{CS Test (Categorical)}} \\ \hline
\multicolumn{1}{l}{DP-CGANS} & \multicolumn{1}{l}{\textbf{0.997}} & \multicolumn{1}{l}{\textbf{0.995}} & \multicolumn{1}{l}{0.982} & \multicolumn{1}{l}{\textbf{0.983}} & \textbf{0.984} \\ 
\multicolumn{1}{l}{CTGAN} & \multicolumn{1}{l}{0.988} & \multicolumn{1}{l}{0.989} & \multicolumn{1}{l}{\textbf{0.984}} & \multicolumn{1}{l}{0.960} & 0.967 \\ 
\multicolumn{1}{l}{MedGAN} & \multicolumn{1}{l}{0.976} & \multicolumn{1}{l}{0.975} & \multicolumn{1}{l}{0.968} & \multicolumn{1}{l}{0.949} & 0.839 \\ 
\multicolumn{1}{l}{TableGAN} & \multicolumn{1}{l}{0.987} & \multicolumn{1}{l}{0.987} & \multicolumn{1}{l}{0.981} & \multicolumn{1}{l}{0.979} & 0.964 \\ \hline
\multicolumn{6}{c}{\textbf{KSTest (Continuous)}} \\ \hline
\multicolumn{1}{l}{DP-CGANS} & \multicolumn{1}{l}{\textbf{0.820}} & \multicolumn{1}{l}{0.796} & \multicolumn{1}{l}{\textbf{0.873}} & \multicolumn{1}{l}{\textbf{0.932}} & \textbf{0.910} \\ 
\multicolumn{1}{l}{CTGAN} & \multicolumn{1}{l}{0.794} & \multicolumn{1}{l}{\textbf{0.819}} & \multicolumn{1}{l}{0.870} & \multicolumn{1}{l}{0.889} & 0.896 \\ 
\multicolumn{1}{l}{MedGAN} & \multicolumn{1}{l}{0.127} & \multicolumn{1}{l}{0.199} & \multicolumn{1}{l}{0.442} & \multicolumn{1}{l}{0.176} & 0.314 \\ 
\multicolumn{1}{l}{TableGAN} & \multicolumn{1}{l}{0.627} & \multicolumn{1}{l}{0.440} & \multicolumn{1}{l}{0.540} & \multicolumn{1}{l}{0.925} & 0.896 \\ \hline
\multicolumn{6}{c}{\textbf{Cramer's V (Dependency between categorical variables)}} \\ \hline
\multicolumn{1}{l}{DP-CGANS} & \multicolumn{1}{l}{0.017} & \multicolumn{1}{l}{\textbf{0.024}} & \multicolumn{1}{l}{\textbf{0.068}} & \multicolumn{1}{l}{0.018} & \textbf{0.029} \\ 
\multicolumn{1}{l}{CTGAN} & \multicolumn{1}{l}{\textbf{0.014}} & \multicolumn{1}{l}{0.031} & \multicolumn{1}{l}{0.085} & \multicolumn{1}{l}{0.030} & 0.041 \\ 
\multicolumn{1}{l}{MedGAN} & \multicolumn{1}{l}{0.061} & \multicolumn{1}{l}{0.130} & \multicolumn{1}{l}{0.148} & \multicolumn{1}{l}{0.063} & 0.116 \\ 
\multicolumn{1}{l}{TableGAN} & \multicolumn{1}{l}{0.024} & \multicolumn{1}{l}{0.031} & \multicolumn{1}{l}{0.102} & \multicolumn{1}{l}{\textbf{0.011}} & 0.030 \\ \hline
\multicolumn{6}{c}{\textbf{Pearson correlation (Correlation between continuous variables)}} \\ \hline
\multicolumn{1}{l}{DP-CGANS} & \multicolumn{1}{l}{\textbf{0.025}} & \multicolumn{1}{l}{\textbf{0.043}} & \multicolumn{1}{l}{\textbf{0.045}} & \multicolumn{1}{l}{\textbf{0.066}} & \textbf{0.020} \\ 
\multicolumn{1}{l}{CTGAN} & \multicolumn{1}{l}{0.033} & \multicolumn{1}{l}{0.064} & \multicolumn{1}{l}{0.050} & \multicolumn{1}{l}{0.132} & 0.056 \\ 
\multicolumn{1}{l}{MedGAN} & \multicolumn{1}{l}{0.487} & \multicolumn{1}{l}{0.718} & \multicolumn{1}{l}{0.324} & \multicolumn{1}{l}{0.279} & 0.542 \\ 
\multicolumn{1}{l}{TableGAN} & \multicolumn{1}{l}{0.077} & \multicolumn{1}{l}{0.058} & \multicolumn{1}{l}{0.046} & \multicolumn{1}{l}{0.092} & 0.051 \\ 
\bottomrule
\end{longtable}
\end{scriptsize}

All models have a common challenge, which is to accurately maintain the dependencies as strong as in real data. For instance, “\textit{Age}” and “\textit{Retirement}”, or “\textit{Employment status}” (\textit{employed}, \textit{unemployed}) and “\textit{Occupational class}” (\textit{high class}, \textit{intermediate}, \textit{low}, \textit{not working}) have strong dependencies between them. All models capture the dependencies to a different extent. TableGAN simulates the most similar strength of these dependencies to the real data, because the third neural network model in addition to the generator and discriminator in TableGAN captures the dependencies and classifies if the generated data is realistic or not. 

For the continuous variables, the inserted conditional vector in DP-CGANS helps in shaping the multimode distributions of the continuous variables and capturing the correlations. However, we found DP-CGANS suffers from oversampling the number of modes in the distributions and handling the variables with a heavy-tailed probability distribution whose tails are not exponentially bounded. This can be observed from the results of the KL divergence test on continuous variables. KL divergence and KS test are both used to observe the difference of two distributions, but only the KL divergence test shows that DP-CGANS does not have a competitive performance. This is because the differences in the tails of the distributions get amplified in KL divergence but not in the KS test.

\subsection{Machine Learning Performance}
\begin{sidewaysfigure}
    \centering
    \includegraphics[scale=0.16]{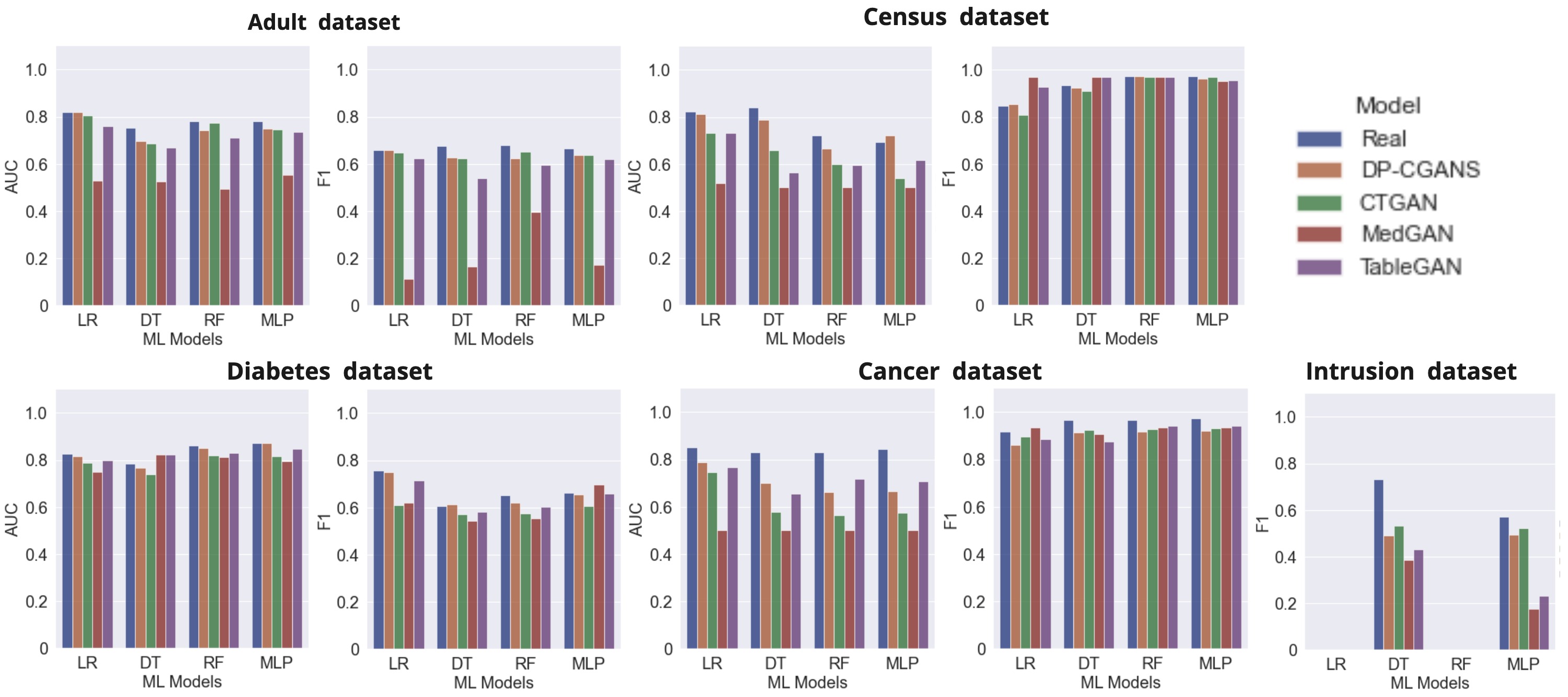}
    \caption{Evaluation results of training machine learning models on synthetic data using different generative models. DT and MLP are conducted to solve the multi-class classification in Intrusion dataset and evaluate the model performance using F1 macro score.}
    \label{cha6fig:mlresults}
\end{sidewaysfigure}

Figure~\ref{cha6fig:mlresults} reports the evaluation results of four machine learning models trained by generated synthetic data. Given that the AUC and F1 score are more reliable in evaluating model performance on the imbalanced datasets, we used these two scores compared with the baseline which results from the real training datasets. The better and more realistic synthetic data is, the smaller the difference in its machine learning performance from the real data. In Adult, Census, and Diabetes, DP-CGANS generates the synthetic data which have the most similar machine learning performance to the real data compared to other models. In these datasets, most variables are imbalanced categorical or binary variables which are handled by inserting the conditional vectors in both DP-CGANS and CTGAN. The advantage of capturing the dependencies between categorical variables in DP-CGANS is reflected on Census and Diabetes. The dependencies of variables in these two datasets have an obvious positive impact on the final classification.

CTGAN shows close performance to DP-CGANS in some experiments and slightly outperformed in the Intrusion dataset. Intrusion dataset has a few extremely imbalanced categorical variables and many continuous variables including heavy-tailed variables. The results show all included models have difficulties generating synthetic data that simulates such extreme distributions from real data. The added conditional vector and extra penalty to the generator strongly encourage DP-CGANS to balance the under-sampling classes and dependencies across categorical variables which weakens the precise mapping of continuous variables. Furthermore, we found TableGAN has comparable performance with CTGAN and DP-CGANS in the Cancer dataset. Cancer dataset has a much smaller size and a simpler structure compared to the other four datasets. As the only one supervised synthetic data generator in the experiment, TableGAN benefits from its third neural network model as an auxiliary classifier and its convolutional GAN structure. Other included models are unsupervised synthetic data generators which typically require more data instances to train.

\subsection{Privacy Cost}

Table~\ref{cha6tab:idresults} reports the privacy costs in \textbf{identity disclosure} with certain threshold distances between synthetic and real data instances. Each dataset applies a different threshold of similarity as the shortest accepted distances between synthetic and real data instances. A lower precision indicates a smaller proportion of “real” data instances labeled by an attacker are presented in the training data. A lower recall indicates fewer real data instances can be detected by an attacker. A low precision and recall achieve a higher level of privacy. MedGAN has the least utility of the synthetic data but holds the greatest privacy guarantee. Note that the privacy score in identity disclosure is under a certain distance threshold (D). For example, MedGAN has no synthetic data instances close to any real data instances with a distance threshold at 0.05. DP-CGANS outperforms other models in Census and Diabetes datasets regarding data utility but has the most privacy costs in identity disclosure. Similar results are observed in Cancer dataset where TableGAN has best utility scores but lowest privacy level. 

%

\begin{table}[htb]
\scriptsize
\centering
\caption{Privacy measurement in identity disclosure on five datasets. Pre represents precision score, while Rec represents recall score. Both scores are 0 to 1.}
\label{cha6tab:idresults}
\begin{tabular}{p{0.13\linewidth}p{0.035\linewidth}p{0.035\linewidth}p{0.035\linewidth}p{0.035\linewidth}p{0.035\linewidth}p{0.035\linewidth}p{0.035\linewidth}p{0.035\linewidth}p{0.035\linewidth}p{0.035\linewidth}}
\hline
\textbf{Dataset} & \multicolumn{2}{c}{\begin{tabular}[c]{@{}c@{}}Adult \\ (D=0.1)\end{tabular}} & \multicolumn{2}{c}{\begin{tabular}[c]{@{}c@{}}Census \\ (D=0.05)\end{tabular}} & \multicolumn{2}{c}{\begin{tabular}[c]{@{}c@{}}Intrusion \\ (D=0.01)\end{tabular}} & \multicolumn{2}{c}{\begin{tabular}[c]{@{}c@{}}Diabetes \\ (D=0.2)\end{tabular}} & \multicolumn{2}{c}{\begin{tabular}[c]{@{}c@{}}Cancer \\ (D=0.2)\end{tabular}} \\ \hline
  & \multicolumn{1}{c}{Pre} & \multicolumn{1}{c}{Rec} & \multicolumn{1}{c}{Pre} & \multicolumn{1}{c}{Rec} & \multicolumn{1}{c}{Pre} & \multicolumn{1}{c}{Rec} & \multicolumn{1}{c}{Pre} & \multicolumn{1}{c}{Rec} & \multicolumn{1}{c}{Pre} & Rec \\ \hline
DPCGANS & \textit{0.518} & \textit{0.254} & \textit{0.576} & \textit{0.193} & \textit{0.489} & \textit{0.389} & \textit{0.495} & \textit{0.188} & \textit{0.585} & 0.088 \\
CTGAN & \textit{0.552} & \textit{0.272} & \textit{0.497} & \textit{0.078} & \textit{0.496} & \textit{0.783} & \textit{0.469} & \textit{0.127} & \textit{0.683} & 0.102 \\
MedGAN & \textit{0} & \textit{0} & \textit{0} & \textit{0} & \textit{0} & \textit{0} & \textit{0} & \textit{0} & \textit{0} & \textit{0} \\
TableGAN & \textit{0.618} & \textit{0.110} & \textit{0.486} & \textit{0.123} & \textit{0} & \textit{0} & \textit{0.488} & \textit{0.179} & \textit{0.895} & 0.162 \\ \hline
\end{tabular}
\end{table}


\begin{table}[]
\centering
\scriptsize
\caption{Privacy measurement of attribute disclosure with using three, six, and all rest known variables. The average scores are reported.}
\label{cha6tab:attriresults}
\begin{tabular}{p{0.13\linewidth}p{0.035\linewidth}p{0.035\linewidth}p{0.035\linewidth}p{0.035\linewidth}p{0.035\linewidth}p{0.035\linewidth}p{0.035\linewidth}p{0.035\linewidth}p{0.035\linewidth}p{0.035\linewidth}}
\hline
\textbf{Dataset} & \multicolumn{2}{c}{Adult} & \multicolumn{2}{c}{Census} & \multicolumn{2}{c}{Intrusion} & \multicolumn{2}{c}{Diabetes} & \multicolumn{2}{c}{Cancer} \\
 & \multicolumn{1}{c}{Cat} & \multicolumn{1}{c}{Con} & \multicolumn{1}{c}{Cat} & \multicolumn{1}{c}{Con} & \multicolumn{1}{c}{Cat} & \multicolumn{1}{c}{Con} & \multicolumn{1}{c}{Cat} & \multicolumn{1}{c}{Con} & \multicolumn{1}{c}{Cat} & Con \\ \hline
\textit{Real} & \textit{0.281} & \textit{0.065} & \textit{0.118} & \textit{0.066} & \textit{0.046} & \textit{0.004} & \textit{0.427} & \textit{0.080} & \textit{0.126} & - \\
DPCGANS & \textit{0.300} & \textit{0.081} & \textit{0.127} & \textit{0.081} & \textit{0.052} & \textit{0.016} & \textit{0.448} & \textit{0.081} & \textit{0.181} & - \\
CTGAN & \textit{0.301} & \textit{0.151} & \textit{0.137} & \textit{0.104} & \textit{0.049} & \textit{0.003} & \textit{0.510} & \textit{0.082} & \textit{0.158} & - \\
MedGAN & \textit{0.372} & \textit{0.266} & \textit{0.361} & \textit{0.263} & \textit{0.319} & \textit{0.196} & \textit{0.501} & \textit{0.216} & \textit{0.258} & \textit{-} \\
TableGAN & \textit{0.305} & \textit{0.090} & \textit{0.138} & \textit{0.142} & \textit{0.057} & \textit{0.036} & \textit{0.492} & \textit{0.084} & \textit{0.137} & - \\ \hline
\end{tabular}
\end{table}

The \textbf{attribute disclosure} measurement was calculated as $1 - P_{attr}$ where $P_{attr}$ is the average posterior probabilities of correctly predicting the unknown (sensitive) variables in real test data. Table~\ref{cha6tab:attriresults} reports the average score of experiments on 3, 6, and all rest known variables to predict the unknown variables. A greater score presents a higher level of privacy. Cancer dataset has only 2 continuous variables which is not sufficient to conduct an evaluation test. Note that the privacy measurement in attribute disclosure is calculated with respect to the real data. This means the probability of predicting unknown variables in the synthetic data is close to (typically higher than) the probability in the real data. DP-CGANS, which generates the most realistic synthetic data among other models, has relatively low privacy levels in the Adult, Census and Diabetes datasets. CTGAN and TableGAN which have better performance in the Intrusion and Cancer datasets respectively have the least privacy guarantees in these datasets.


\begin{figure}[!htp]
    \centering
    \includegraphics[scale=0.4]{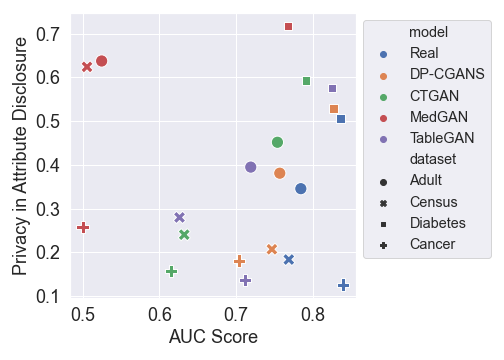}
    \caption{AUC scores and privacy level in attribute disclosure of four generators.}
    \label{cha6fig:aucresults}
\end{figure}


The privacy measurements in both attribute and identity disclosure show the trade-off between synthetic data utility and real data privacy. Figure~\ref{cha6fig:aucresults} plots the AUC scores as the indicator of data utility and privacy level against attribute disclosure as the indicator of data privacy. In all datasets, we found the model that obtains a higher AUC score has a lower level of privacy preservation. On the one hand, we define a generative model is good if it can produce synthetic data as similar as possible to the real data. This means the generator in GAN is motivated to minimize the distance between real and synthetic data. On the other hand, privacy measurement shows a generator outputs data that has a smaller distance to the real data takes more privacy risk in revealing sensitive information from the real data. Therefore, it is an inevitable trade-off between data privacy and data utility in generating synthetic data. Finally, it is found that all included models obtain a relatively low level of privacy in the experiments. It explains the essentiality and necessity of enhancing the privacy guarantee to the construction and training process of generative models.

\subsection{\dpcgans\ with different privacy budget}

\begin{figure}[!htb]
    \centering
    \includegraphics[scale=0.14]{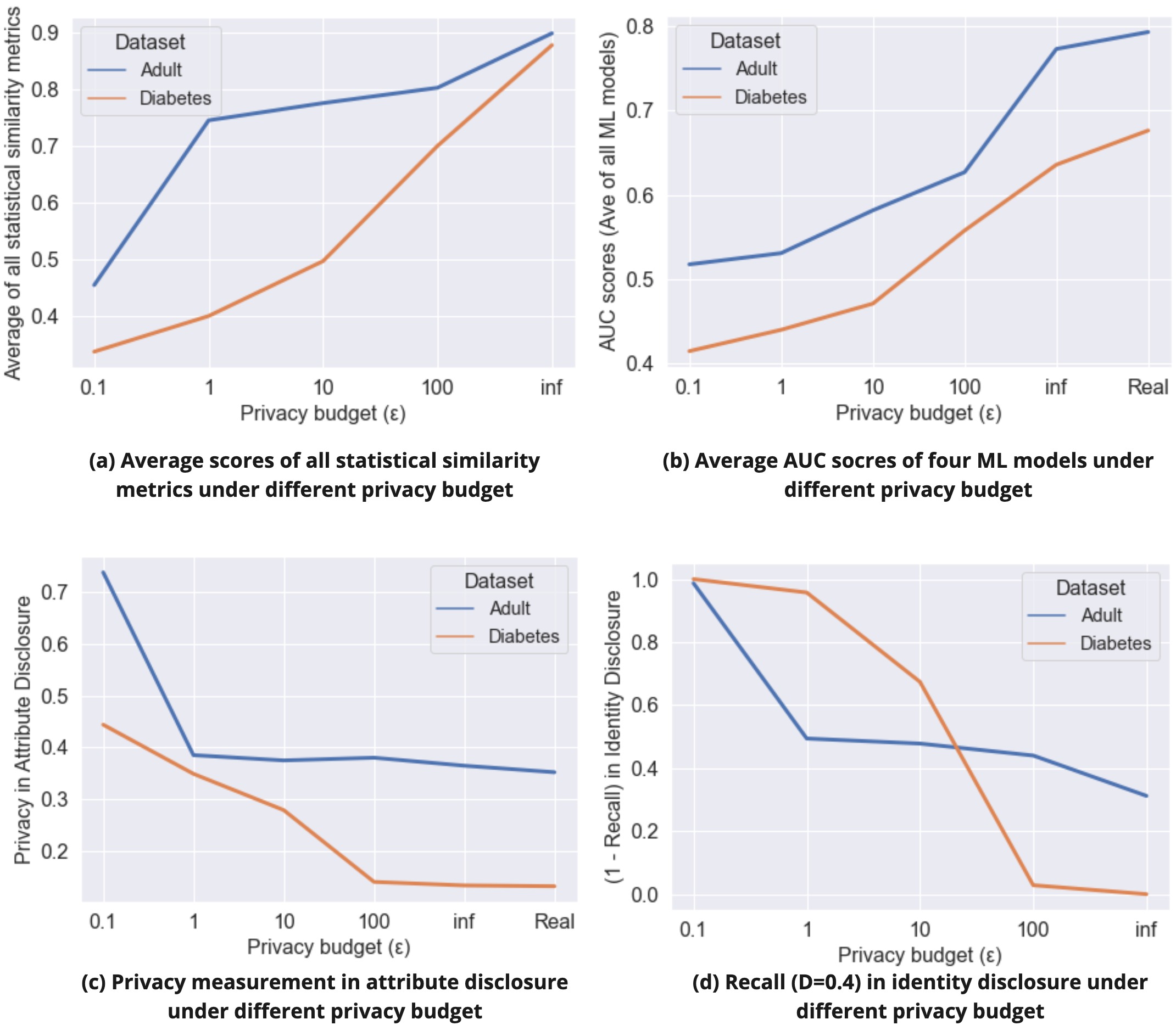}
    \caption{Statistical similarity, ML performance, privacy measurement in attribute and identity disclosure of DP-CGANS under different DP budgets.}
    \label{cha6fig:privacy_identity}
\end{figure}

We experimented a set of the privacy budgets of DP-CGANS ($\varepsilon$ = 0.1, 1, 10, 100, and $\infty$) to enhance the privacy guarantee. Figure~\ref{cha6fig:privacy_identity} shows the overall changes of model performance using different privacy budgets on Adult and Diabetes datasets. The average scores are plotted in the figure, while the full results are presented in Table~\ref{app_tab:privacyresults}. Figure~\ref{cha6fig:privacy_identity}(a) and (b) shows the statistical similarity and ML performance are climbing up as the privacy budget ($\varepsilon$) increases. A larger privacy budget indicates a smaller scale of noises are added into the model training process which means a lower the level of privacy is preserved in the synthetic data. This is proven by the privacy measurement in attribute and identity disclosure under increasing privacy budget as Figure~\ref{cha6fig:privacy_identity}(c) and \ref{cha6fig:privacy_identity}(d) show.


Both datasets demonstrate the trade-off between model performance and privacy level with similar overall changes under increasing privacy budgets. However, the privacy budgets show different impacts on the learning performance of the model on different datasets. When adding $\varepsilon$ from 0.1 to 1, the model has an obvious improvement in statistical similarity on the Adult dataset (figure~\ref{cha6fig:privacy_identity}(a)). When $\varepsilon > 1$, this increase becomes slower. The corresponding changes are observed in figure~\ref{cha6fig:privacy_identity}(c) and \ref{cha6fig:privacy_identity}(d) that the level of privacy drops steeply when $\epsilon$ increases from 0.1 to 1. The model has a relatively stable increase on the Diabetes dataset with a turning point at $\epsilon$=10. Figure~\ref{cha6fig:privacy_identity}(d) shows the impact of privacy budget on the model performance becomes dramatic when $\epsilon$ is between 10 and 100. Although the Adult and Diabetes datasets are both imbalanced datasets and have the same ratio of categorical and continuous variables, the model reacts on the same privacy budget with different sensitivities in different datasets. Obtaining the most optimal balance between model performance and privacy guarantee depends on the data structure and characteristics of each dataset (such as imbalance variables, abnormal distributions, sparsity of data).

\subsection{Limitations}
The empirical results in this study should be considered in the light of limitations. First, in the experiments, we sampled 25\% of data instances from the Census and Intrusion datasets in a stratified random manner owing to the computation resources and time. Using subsamples of the data may limit the performance of the generative models. However, given the proportion of our samples from the original dataset and the model performance reported by other comparable studies, we do not expect that the sampled data would have a significant impact on the final experiments results. 

Second, the conditional vector in the generator of DP-CGANS can successfully capture the dependencies between each pair of variables. However, the generator does not learn to maintain the relations among more than two variables. Extending the construction of the conditional vector to three or more variables dramatically increases the dimension of the vector  at the expense of training efficiency. A potential solution can be training the generative model in a semi-supervised or supervised manner such as selectively including the variables and categories in the conditional vectors or introducing a classifier which is trained with the generator. Further research can be conducted in this direction to improve the capture of dependencies among multiple variables. 


DP-CGANS can output differential private (DP) synthetic data with imbalanced variables and keeping the dependencies between variables mainly because of the generator structure. However, the differential privacy budget (noise) is added to the discriminator and indirectly affects the generator. The data utility of the synthetic data might drop dramatically when tuning the differential privacy budget. Therefore, to obtain the optimal balance between data utility and privacy, DP-CGANS costs computation, time, and effort to carefully find the best suitable privacy budget for each dataset. In the future work, we intend to control the impact of adding DP to the network training on the generator. The solution could be to apply DP directly on the generator instead of discriminator or stabilize the effect of adding noise on the loss which the generator receives from the discriminator.  

\section{Conclusion}
We proposed DP-CGANS, a differentially private conditional GAN,  consisting of four main components - transforming, sampling, conditioning, and network training to generate realistic and privacy-preserving synthetic data. DP-CGANS handles data with mixed  types and imbalanced variables and captures the correlations and dependencies between variables with privacy guaranteed. We compared our model with state-of-the-art generative models on three public datasets and two real-world personal health datasets using a set of extensive evaluation matrices focusing on the statistical similarity, machine learning performance, and privacy measurement. The evaluation results show that our model outperforms other comparable models, especially in capturing dependency between variables. Meanwhile, we measured the privacy risks in different generative models regarding attribute disclosure and identity disclosure. Our experiments prove the trade-off between output data utility (synthetic data) and input data privacy (real data) and our model can reduce privacy risks to a certain extent while maintaining data quality. Finally, we discussed the limitations of DP-CGANS and provided future directions to improve the generation of synthetic data using the developed framework.


\section*{Acknowledgement}
This project is funded by the Dutch Open Data Infrastructure for Social Science and Economic Innovations (ODISSEI). This research was made possible, in part, using the Data Science Research Infrastructure (DSRI) hosted at Maastricht University. Special thanks to Vincent Emonent for helping setting up the GPU environment for experiments. Thanks to Mirela Popa and Thales Bertaglia for their valuable comments to improve the manuscript. 

\bibliography{references}

\clearpage

\appendix
\section{Details of datasets and models parameters}
\label{appendixA}
The characteristics and accessibility of the each dataset used by this study is listed in Table~\ref{cha6tab:datatable}. The public datasets are described and can be accessed from the links in the table. The two real-world health datasets can be requested for research purposes. The Diabetes dataset was collected by The Maastricht Study, an observational prospective population-based cohort study. The rationale and methodology have been described previously~\cite{schram2014maastricht}. Eligible for participation were all individuals aged between 40 and 75 years living in the southern part of the Netherlands. The study has been approved by the institutional medical ethical committee (NL31329.068.10) and the Minister of Health, Welfare and Sports of the Netherlands (Permit 131088-105234-PG). All participants gave written informed consent. The Lung Cancer dataset, which was used to build a prognostic radiomic signature, was collected by Maastro Clinic. The details and methods are described in ~\cite{aerts_data_2019}. The users are required to abide by the Creative Commons Attribution-NonCommercial 3.0 Unported License under which it has been published. 
\clearpage

\section{Additional Results}

\setlength{\tabcolsep}{3pt}
\begin{table}[!htb]
\scriptsize
\centering
\caption{Full results of machine learning model performance using synthetic data generated from different generations models. Logistic regression (LR), decision tree (DT), random forest (RF), and multilayer perceptron (MLP) models are included in the experiments.}
\label{app_tab:MLresults}
\begin{tabular}{llrrrrccrrrr}
\hline
 &  & \multicolumn{2}{c}{Adult} & \multicolumn{2}{c}{Census} & \multicolumn{2}{c}{Intrusion} & \multicolumn{2}{c}{Diabetes} & \multicolumn{2}{c}{Cancer} \\
 &  & \multicolumn{1}{c}{AUC} & \multicolumn{1}{c}{F1} & \multicolumn{1}{c}{AUC} & \multicolumn{1}{c}{F1} & AUC & F1 & \multicolumn{1}{c}{AUC} & \multicolumn{1}{c}{F1} & \multicolumn{1}{c}{AUC} & \multicolumn{1}{c}{F1} \\ \hline
\multirow{5}{*}{LR} & \textbf{Real} & \textit{0.820} & \textit{0.660} & \textit{0.822} & \textit{0.846} & - & - & \textit{0.828} & \textit{0.755} & \textit{0.852} & \textit{0.916} \\
 & DP-CGANS & \textbf{0.818} & \textbf{0.658} & \textbf{0.811} & \textbf{0.853} & - & - & \textbf{0.815} & \textbf{0.751} & \textbf{0.787} & 0.862 \\
 & CTGAN & 0.807 & 0.647 & 0.732 & 0.809 & - & - & 0.787 & 0.610 & 0.745 & \textbf{0.897} \\
 & MedGAN & 0.528 & 0.112 & 0.520 & 0.971 & - & - & 0.748 & 0.622 & 0.500 & 0.936 \\
 & TableGAN & 0.761 & 0.625 & 0.730 & 0.929 & - & - & 0.799 & 0.714 & 0.767 & 0.887 \\ \hline
\multirow{5}{*}{DT} & \textbf{Real} & \textit{0.754} & \textit{0.676} & \textit{0.840} & \textit{0.935} & \textit{-} & \multicolumn{1}{r}{\textit{0.731}} & \textit{0.784} & \textit{0.605} & \textit{0.830} & \textit{0.968} \\
 & DP-CGANS & \textbf{0.696} & \textbf{0.626} & \textbf{0.789} & \textbf{0.924} & - & \multicolumn{1}{r}{0.491} & \textbf{0.768} & \textbf{0.612} & \textbf{0.700} & 0.915 \\
 & CTGAN & 0.687 & 0.624 & 0.659 & 0.911 & - & \multicolumn{1}{r}{\textbf{0.533}} & 0.740 & 0.571 & 0.577 & \textbf{0.923} \\
 & MedGAN & 0.525 & 0.166 & 0.503 & 0.970 & - & \multicolumn{1}{r}{0.385} & 0.824 & 0.544 & 0.500 & 0.906 \\
 & TableGAN & 0.667 & 0.539 & 0.562 & 0.969 & - & \multicolumn{1}{r}{0.432} & 0.821 & 0.582 & 0.656 & 0.876 \\ \hline
\multirow{5}{*}{RF} & \textbf{Real} & \textit{0.782} & \textit{0.679} & \textit{0.720} & \textit{0.973} & \textit{-} & \textit{-} & \textit{0.861} & \textit{0.653} & \textit{0.830} & \textit{0.968} \\
 & DP-CGANS & 0.763 & 0.635 & \textbf{0.665} & \textbf{0.973} & - & - & \textbf{0.850} & \textbf{0.621} & 0.661 & 0.916 \\
 & CTGAN & \textbf{0.775} & \textbf{0.651} & 0.598 & 0.970 & - & - & 0.819 & 0.575 & 0.562 & 0.928 \\
 & MedGAN & 0.494 & 0.395 & 0.500 & 0.971 & - & - & 0.814 & 0.552 & 0.500 & 0.936 \\
 & TableGAN & 0.712 & 0.597 & 0.595 & 0.970 & - & - & 0.830 & 0.603 & \textbf{0.717} & \textbf{0.943} \\ \hline
\multirow{5}{*}{MLP} & \textbf{Real} & \textit{0.780} & \textit{0.664} & \textit{0.692} & \textit{0.972} & \textit{-} & \multicolumn{1}{r}{\textit{0.570}} & \textit{0.871} & \textit{0.662} & \textit{0.845} & \textit{0.972} \\
 & DP-CGANS & \textbf{0.750} & \textbf{0.636} & \textbf{0.721} & 0.964 & - & \multicolumn{1}{r}{0.493} & \textbf{0.873} & 0.656 & 0.666 & 0.921 \\
 & CTGAN & 0.746 & 0.636 & 0.540 & \textbf{0.970} & - & \multicolumn{1}{r}{\textbf{0.522}} & 0.817 & 0.605 & 0.575 & 0.931 \\
 & MedGAN & 0.553 & 0.172 & 0.500 & 0.951 & - & \multicolumn{1}{r}{0.177} & 0.793 & 0.699 & 0.500 & 0.936 \\
 & TableGAN & 0.736 & 0.619 & 0.617 & 0.956 & - & \multicolumn{1}{r}{0.231} & 0.848 & \textbf{0.660} & \textbf{0.707} & \textbf{0.943} \\ \hline
\end{tabular}
\end{table}

\begin{table}[!htb]
\scriptsize
\centering
\caption{Full results of DP-CGANS's performance in statistical similarity, machine learning performance, and privacy costs in attribute and identity disclosure using different privacy budgets.}
\label{app_tab:privacyresults}
\begin{tabular}{ll|rrrrrl}
\hline
\multicolumn{2}{l}{\multirow{2}{*}{}} & \multicolumn{6}{c}{\textbf{Adult}} \\
\multicolumn{2}{l}{} & \multicolumn{1}{l}{$\varepsilon$=0.1} & \multicolumn{1}{l}{$\varepsilon$=1} & \multicolumn{1}{l}{$\varepsilon$=10} & \multicolumn{1}{l}{$\varepsilon$=100} & \multicolumn{1}{l}{$\varepsilon$=$\infty$} & Real \\ \hline
\multirow{4}{*}{\begin{tabular}[c]{@{}l@{}}Data\\ Privacy\end{tabular}} & AD-Cat & 0.602 & 0.282 & 0.290 & 0.300 & 0.292 & \multicolumn{1}{r}{0.281} \\
 & AD-Con & 0.137 & 0.103 & 0.084 & 0.081 & 0.073 & \multicolumn{1}{r}{0.071} \\
 & ID-Pre & 0.371 & 0.479 & 0.479 & 0.495 & 0.493 & - \\
 & ID-Rec & 0.013 & 0.506 & 0.522 & 0.560 & 0.828 & - \\ \hline
\multirow{8}{*}{\begin{tabular}[c]{@{}l@{}}Data \\ Utility\end{tabular}} & AUC & 0.518 & 0.531 & 0.582 & 0.627 & 0.773 & \multicolumn{1}{r}{0.793} \\
 & F1 & 0.090 & 0.232 & 0.287 & 0.329 & 0.667 & \multicolumn{1}{r}{0.683} \\
 & KLC & 0.325 & 0.687 & 0.661 & 0.722 & 0.889 & - \\
 & KLD & 0.541 & 0.833 & 0.817 & 0.802 & 0.935 & - \\
 & CS & 0.812 & 0.991 & 0.981 & 0.974 & 0.991 & - \\
 & KS & 0.140 & 0.471 & 0.645 & 0.715 & 0.783 & - \\
 & CV & 0.203 & 0.142 & 0.139 & 0.145 & 0.029 & - \\
 & CORR & 0.067 & 0.066 & 0.077 & 0.068 & 0.031 & - \\ \hline
\end{tabular}
\begin{tabular}{ll|rrrrrl}
\hline
\multicolumn{2}{l|}{\multirow{2}{*}{}} & \multicolumn{6}{c}{\textbf{Diabetes}} \\
\multicolumn{2}{l|}{} & \multicolumn{1}{l}{$\varepsilon$=0.1} & \multicolumn{1}{l}{$\varepsilon$=1} & \multicolumn{1}{l}{$\varepsilon$=10} & \multicolumn{1}{l}{$\varepsilon$=100} & \multicolumn{1}{l}{$\varepsilon$=$\infty$} & Real \\ \hline
\multirow{4}{*}{\begin{tabular}[c]{@{}l@{}}Data\\ Privacy\end{tabular}} & AD-Cat & 0.314 & 0.237 & 0.228 & 0.090 & 0.083 & \multicolumn{1}{r}{0.080} \\
 & AD-Con & 0.129 & 0.112 & 0.050 & 0.050 & 0.050 & \multicolumn{1}{r}{0.051} \\
 & ID-Pre & 0.000 & 0.276 & 0.405 & 0.499 & 0.501 & - \\
 & ID-Rec & 0.000 & 0.043 & 0.326 & 0.972 & 1.000 & - \\ \hline
\multirow{8}{*}{\begin{tabular}[c]{@{}l@{}}Data \\ Utility\end{tabular}} & AUC & 0.415 & 0.440 & 0.471 & 0.558 & 0.636 & \multicolumn{1}{r}{0.676} \\
 & F1 & 0.146 & 0.195 & 0.217 & 0.220 & 0.444 & \multicolumn{1}{r}{0.512} \\
 & KLC & 0.173 & 0.180 & 0.181 & 0.441 & 0.647 & - \\
 & KLD & 0.581 & 0.677 & 0.792 & 0.798 & 0.972 & - \\
 & CS  & 0.549 & 0.700 & 0.861 & 0.855 & 0.988 & - \\
 & KS  & 0.036 & 0.048 & 0.153 & 0.707 & 0.908 & - \\
 & CV  & 0.068 & 0.071 & 0.048 & 0.047 & 0.057 & - \\
 & CORR & 0.457 & 0.526 & 0.504 & 0.543 & 0.342 & - \\ \hline
\end{tabular}
\end{table}

\end{document}